\documentclass[runningheads]{llncs}
\usepackage{graphicx}
\usepackage[title]{appendix}
\begin{document}
\title{Don't Take it Personally: Analyzing Gender and Age Differences in Ratings of Online Humor
}
%
\titlerunning{Analyzing Gender and Age Differences in Ratings of Online Humor}
%
\author{J.A. Meaney\inst{1}\orcidID{0000-0002-7624-4429} \and
Steven R. Wilson \inst{1,2}\orcidID{0000-0002-2458-0439} \and
Luis Chiruzzo \inst{3}\orcidID{0000-0002-1697-4614} \and
Walid Magdy \inst{1}\orcidID{0000-0001-9676-1338}}

\authorrunning{J.A. Meaney et al.}
%
\institute{School of Informatics, University of Edinburgh, UK \\
\email{jameaney@ed.ac.uk, wmagdy@inf.ed.ac.uk}\and
Oakland University, Rochester, Michigan, US \\
\email{stevenwilson@oakland.edu}\\ \and
Universidad de la República, Montevideo, Uruguay\\
\email{luischir@fing.edu.uy}}
\maketitle              
\begin{abstract}
Computational humor detection systems rarely model the subjectivity of humor responses, or consider alternative reactions to humor - namely offense. We analyzed a large dataset of humor and offense ratings by male and female annotators of different age groups. We find that women link these two concepts more strongly than men, and they tend to give lower humor ratings and higher offense scores. We also find that the correlation between humor and offense increases with age. Although there were no gender or age differences in humor detection, women and older annotators signalled that they did not understand joke texts more often than men. We discuss implications for computational humor detection and downstream tasks.  

\keywords{Computational Humor  \and Offense Detection \and Online Texts \and Demographics.}
\end{abstract}
\section{Introduction}
Computational Humor Detection is a fast-growing area of research and has produced at least one humor detection challenge per year since 2017 with Hashtag Wars in SemEval 2017, ~\cite{potash2017semeval}, the Spanish-language HAHA task in Iberlef 2018~\cite{castro2018overview} and 2019~\cite{chiruzzo2019overview}, Assessing Humor in Edited Headlines in 2020~\cite{hossain2020semeval} and HaHackathon in 2021~\cite{meaney2021semeval}. With participation in these challenges increasing year on year, organisers are beginning to refine their conception of humor, and to incorporate some of the vast, inter-disciplinary findings of the broader humor research community. 

One vital branch of this research is that humor is known to vary along the lines of demographic characteristics. Factors such as age~\cite{kuipers2015humor}, gender~\cite{hofmann2020gender}, personality~\cite{ruch2010sense} and other demographic variables all modulate responses to humor. Humor tasks have struggled to incorporate such demographic awareness into their tasks, and instead tend to average over all humor ratings - which removes nuance and subjectivity from the data~\cite{meaney2020crossing}, as well as possibly decreasing the generalizability of humor detection systems. 

A second salient finding from the broader humor literature is that humor is closely linked to offense~\cite{lockyer2005beyond} and indeed, can be used as a mechanism to mask hateful or offensive content. Several competitions have modelled hate speech~\cite{basile2019semeval}~\cite{zampieri2020semeval}, which is related to offense, but HaHackathon was the first humor detection competition to co-model humor and offense. As the concept of offense is less tangible than humor, it was split in two:
\begin{enumerate}
    \item \textit{General offense} meaning that a text targets a group of people simply because they belonged to that group and/or is likely upsetting to a lot of people.
    \item \textit{Personal offense}, targeting a group that the reader belongs to or cares about.
\end{enumerate}
 Although the annotators of this dataset provided demographic data about their age and gender, this was not released as part of the humor detection task, and this is the first analysis of the impact of these age and gender on the humor and offense ratings in this large dataset. The analysis aims to uncover if humor and offense are as meaningfully linked in big datasets as they are in small-N studies, while validating evidence that there are gendered differences in the distribution of humor ratings~\cite{svebak2004prevalence}, as well as tolerance of aggressive humor. 
 
 As in \cite{hofmann2020gender}, we are mindful of the use of \textit{gender} to specify a cultural phenomenon, indicating men and women as socially-defined groups, rather than a biological distinction. 

\subsection{Related Work}
Gender and age differences have been the subject of many studies in the fields of psychology, sociology, education, and management studies. Svebak et al. \cite{svebak2004prevalence} found that “overall humor scores” were higher for men than they were for women. However, it should be noted that “overall humor” was narrowly assessed, using only three items, with each representing one of the dimensions of the Situational Humor Questionnaire. The same work reported that humor appreciation declines with age: the mean scores for total sense of humor on average declined across the age cohorts from highest score in the 20s to lowest score among those aged 70.  More recently, an Italian study of covid-related humor \cite{bischetti2021funny} reported that increasing age, as well as being female was related to finding pandemic humor more aversive and less funny. 

In terms of gender differences, perhaps the most replicated result is that men tolerate aggressive humor more than female respondents do \cite{hofmann2020gender}. Proyer and Ruch \cite{proyer2010enjoying} report that men tended to score higher on kagelaticism - the joy of laughing at others, which suggests that as long as a joke does not target men explicitly, it may be offensive towards other groups, without impacting men’s humor ratings. Interestingly, Knegtmans et al. \cite{knegtmans2018impact} found that participants whose social power had been manipulated to place them in a high-power state rated jokes which targeted others as less offensive, and gave higher humor ratings. No differences in the appreciation of nonsense humor \cite{kohler1996sources}, or neutral jokes \cite{ferstl2017humor} were found.

\subsection{Research Questions (RQ)}
\begin{enumerate}
    \item Is there a correlation between annotators' perceptions of humor and offense? Does this vary by age and gender?
    \item Are there differences in humor \textit{detection} and \textit{comprehension} between groups?
    \item Are there differences in the distributions of humor and offense ratings between groups?
\end{enumerate}

Using a dataset of $>$120k ratings of humor and offense~\cite{meaney2021semeval}, we find a slight negative correlation between humor and offense, which varies as a function of gender and age. The negative link between humor and offense increases as annotators age. We also find a stronger correlation between general offense and humor for women, but male annotators only linked these concepts when they signalled that they were personally offended. There were no significant differences between groups when it came to correctly identifying texts as jokes (i.e. humor detection), but there were differences when it came to humor comprehension. More women than men indicated that they did not get a joke, and women of all age groups had higher rates of using the label ``I don't get it'' than men of all age groups. In terms of the distribution of ratings, women were more likely to use lower humor ratings and higher offense ratings, while men showed the opposite trend. In terms of age groups, the oldest group tended to report that they didn't get a joke more than any other group, while annotators ages 26-40 were least likely to use this label, and also gave the highest humor ratings overall. Older groups were more likely to use higher ratings of general and personal offense, while younger annotators were less likely to use these.

\section{Dataset Description}

The dataset features the texts and ratings used in the humor and offense shared task HaHackathon at SemEval 2021~\cite{meaney2021semeval}. Including non-humorous texts, this comprises 202,369 ratings of 10,000 texts. Each text has an average of 20.2 ratings, with no text having fewer than 17 votes. There were 1,821 unique annotators (mean age 40.45 years, SD=15.64 years), and each annotator rated an average of 111.13 texts. The highest number of texts rated by one person was 307.

Of the 10,000 texts in the dataset, 2,000 were sourced from the Kaggle Short Jokes Dataset\footnote{https://github.com/amoudgl/short-jokes-dataset}. Half of the Kaggle texts were selected because they referred to one of the common targets of online hate speech outlined by Silva et al. ~\cite{silva2016analyzing}, e.g. women, members of the LGBT community, religious/racial minorities, and this target was the butt of the joke. These texts were deemed likely to elicit ratings of offense from some annotators. The other half of the Kaggle texts referred to a common hate speech target, but did not make it the butt of the joke. 

The other 8,000 texts were sourced from Twitter, from a mix of humorous and non-humorous accounts. Amongst the non-humorous accounts, there were several which advocate for, or provide information to common targets of hate speech. This ensured that mentions of these targets were not limited to humorous texts only. 

Annotators were asked up to three sets of questions about each text: one related to humor and two related to offense. 

\begin{enumerate}
    \item \textbf{Humor detection/rating:} annotators were asked if the intention of the text was to be humorous. This binary response question was aimed at gauging the genre of the text, and annotators were asked not to judge based on whether they found it funny, but whether it contained indicators of the humor genre, e.g. a setup and punchline, puns, absurd content, etc. If the annotator selected ‘yes’, they were asked to rate how funny they found it from 1-5. There was also the option to select ‘I don’t get it’ if the text was identified as humorous, but the humor was not understood. If the annotator selected ‘no’, they were not asked any further questions about this text. 
    \item \textbf{General offense detection/rating:} If a text had been labelled as humorous, annotators were asked if they thought the text targeted a group simply because they belonged to a group, or if they thought the text would be offensive to a large number of people. In the case of a ‘yes’ response, they were asked how generally offensive they thought the text was from 1-5.
    \item \textbf{Personal offense:} If a text had been labelled as humorous, we asked annotators if they were personally hurt by the text, or were hurt on someone else’s behalf, and if so, to rate how much from 1-5.

\end{enumerate}

The pool of annotators comprised 4 age groups: 18-25, 26-40, 41-55, 56-70. In order to avoid a lack of shared cultural knowledge, all annotators were native English speakers and citizens of the United States. Although we aimed to be inclusive of diverse genders, the dataset included only four annotators who preferred not to disclose their gender. As they rated a total of 384 texts, they were excluded from the gender analysis, for reasons of statistical power. 

Annotators provided informed consent before beginning the annotation, and the procedure was approved by the Ethics Committee of the corresponding author's institution. Other demographic data about the annotators, such as gender and personality traits, was also provided as part of the dataset.

\section{Methodology}

Given that the humor and offense annotations were reported using an ordinal scale, for RQ1, we used the Spearman rank correlation \cite{spearman1904american} to report the correlations between these variables. The Spearman rank correlation is a generalisation of the Pearson correlation which is used for discrete and ordinal data which captures the strength and direction of the relationship between two variables by ranking the values of each variable, summing the square differences and calculating the covariance of the ranks. This returns a correlation coefficient, $\rho$, ranging from -1 to +1, the magnitude of which indicates the strength of the relationship and the sign signifies the direction. It also returns a \textit{p}-value - the probability that the value of the coefficient could occur under the null hypothesis. 

To answer RQ2, we calculated the proportion of annotators from each group (i.e., gender or age group) that mislabeled (failed to \textit{detect}) or misunderstood (failed to \textit{comprehend}) each text. The resulting distributions were non-normal, so we chose non-parametric tests, which do not assume an underlying distribution. As we have only two values for gender in the dataset, we used a Wilcoxon Signed Rank test \cite{wilcoxon1945individual} to examine the null hypothesis that the samples from male and female annotators came from the same distribution. This is similar to a paired t-test, and it ranks the absolute value of the pairs of differences to calculate the test statistic, \textit{w}. With this test, we report the Common Language Effect Size (CLES): the proportion of pairs where the values for one group are higher than the other. 

For more than two groups, i.e., our age variable, which had four bins, we use the Friedman test \cite{friedman1937use}, which is similar to a repeated measures ANOVA. Again values are ranked and the test compares the mean rank of each group for statistical significance. In the case of a significant result, we ran post hoc pairwise Wilcoxon tests. We used the Bonferroni correction to adjust the p-values for multiple comparisons, reducing the risk of false positive results. 

For RQ3, we first used the Wilcoxon and Friedman tests to determine if one group tended to give higher or lower ratings than another. We then used a chi-square test of homogeneity to examine how the distributions differed from each other. This test determines if the frequencies of each possible value of the dependent variable are distributed in the same way across the different groups. The test calculates the expected frequencies of each rating by each group by multiplying the number of annotators in each group by the true probability that any annotator would pick each answer. This expected frequency is then compared to the observed frequency. 

\section{Results}

\subsection{RQ1: Is There a Correlation between Humor and Offense?}

For the following analysis, we excluded texts which had been labelled as ‘not humorous’ by our annotators, and removed outliers (e.g. texts that had fewer than 3 humor ratings). This left 121,622 ratings of 6,918 texts.

Overall, there was a small negative correlation between humor and general offense ($\rho=-0.13$, $p<$0.05), and this grew stronger for humor and personal offense ($\rho$ =-0.19, $p <$0.05), which suggests that offensive content is negatively related to humor appreciation. There was a strong correlation between general and personal offense ($\rho$=0.60, $p <$0.05), indicating that these concepts are linked, but are not identical. 

\begin{figure}[htp!]
\caption{Correlations between Humor and Offense by Gender and Age}
\centering
\includegraphics[width=1.0\textwidth]{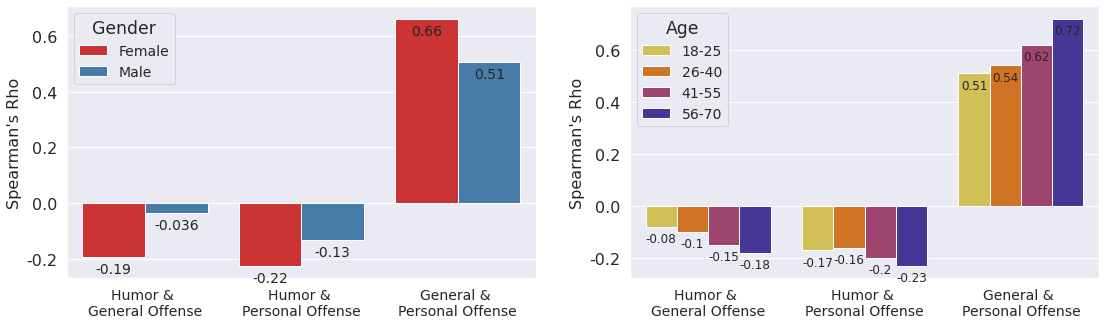}
\label{fig:correlations_by_gender_and_age}
\end{figure}
\subsubsection{Correlations between ratings by Gender}
When examining the correlations between ratings split by gender, an interesting trend emerged (Figure \ref{fig:correlations_by_gender_and_age}). There was almost no relationship between humor and \textit{general} offense for men, however \textit{personal} offense ratings were negatively correlated with humor ratings. Conversely, for female annotators, both types of offense were more strongly correlated with a reduced humor rating for female annotators.

\subsubsection{Correlations by Age}

A second interesting trend emerged in terms of age: the older the annotators were, the stronger the negative link between general \textit{and} personal offense on humor ratings was (Figure \ref{fig:correlations_by_gender_and_age}). The oldest group had the most prominent negative correlation between humor and both types of offense, as well as the strongest correlation between the two offense metrics.

\subsubsection{Correlations by Age and Gender}

\begin{figure}[h!]
\caption{Correlations between Humor and Offense by Age and Gender}
\centering
\includegraphics[width=0.9\textwidth]{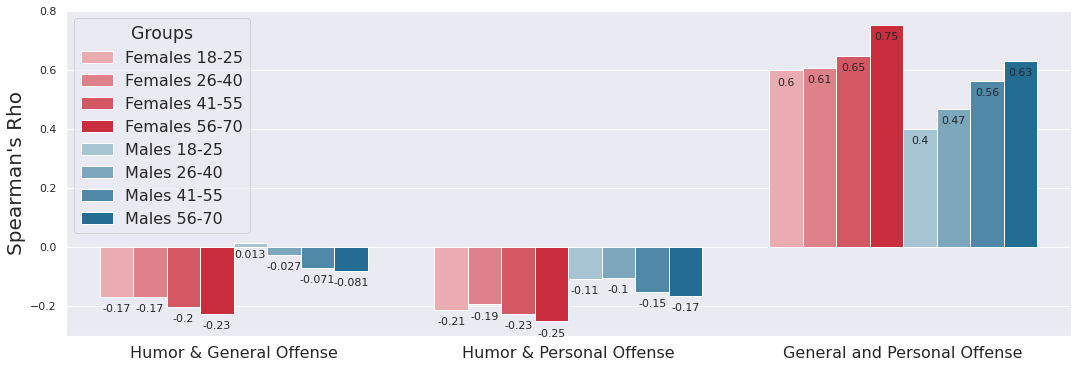}
\label{fig:gender-age-corr}
\end{figure}

Although splitting 20 ratings per text into 8 groups (for four age groups by two gender groups) would cause issues of data sparsity and statistical power, we noted that the trend of an increasingly negative correlation between humor and offense continues when this is broken down by age and gender (Figure \ref{fig:gender-age-corr}). Female annotators relate lower humor scores to higher offense scores increasingly with age, and this trend is much less pronounced in male annotators. 

\subsection{RQ2: Are There Differences in Humor Detection and Comprehension Between Groups?}

\subsubsection{Humor Detection}

To investigate differences in annotators' humor \textit{detection}, we looked at the proportion of male and female annotators who labelled each text from the Kaggle data as ‘not humorous’. We confined this analysis to the Kaggle data because all texts in this dataset was intended to be humorous, and should have been labeled as such. A paired Wilcoxon signed rank test showed that there was no significant difference between groups ($z$ = 134201.0, $p$=0.29). 

\begingroup
\setlength{\tabcolsep}{10pt} 
\renewcommand{\arraystretch}{1}
\begin{table}[h!]
\caption{Mislabeling and Misunderstanding in the Kaggle Jokes}
\label{tab:mislabeled-jokes}
\begin{tabular}{lll}
\hline
 & \textbf{Male} & \textbf{Female} \\ \hline
Proportion of annotations from each group                          & 42.92\%        & 57.08\%          \\ \hline
'Not Humor' ratings from each group                      & 3.79\%        & 3.70\%           \\ \hline
Unique texts with 1+ label of not-humorous       & 22.91\%          & 25.42\%            \\ \hline
‘I don’t get it’ ratings from each group         & 5.77\%         & 7.35\%           \\ \hline
Unique texts with 1+ rating of ‘I don’t get it’’ & 33.04\%          & 45.81\%           \\ \hline
\end{tabular}
\end{table}
\endgroup

We used a similar procedure to test if there were significant differences between age groups in terms of humor detection. A Friedman test showed that there were no significant differences between groups ($\chi^2$ = 6.976, $p$=0.07).

\subsubsection{Humor Comprehension}

After labeling a text as humorous, one of the options for humor rating was ‘I don’t get it’. This indicated that the annotator had recognized that the text was intended to be humorous, but that they lacked the knowledge to fully understand the joke. We first looked at the Kaggle dataset, and calculated the number of ‘I don’t get it’ votes from men and women, as a proportion of the total votes per text from each group. A paired Wilcoxon signed rank test showed that there was a significant difference between groups ($z$ = 214403.0, $p<$0.05). We used Pingouin~\cite{vallat2018pingouin} to calculate the Common Language Effect Size (CLES), i.e. the proportion of pairs where the proportion of ‘I don’t get it’ ratings provided by female annotators is greater than the proportion of male annotators who gave that rating. The resulting CLES of 0.5540 indicates that a larger proportion of female annotators indicated that they did not get the joke in 55.45\% of pairs. When looking at the data from Twitter, women still admit to not getting the joke more than men ($z$ = 2298680.0, $p<$0.05), but the effect is less pronounced, CLES = 0.5223. 

We examined differences between age groups in terms of humor detection. A Friedman test showed that there were no significant differences between groups ($\chi^2$= 0.0012, $p$=0.06).

\subsection{RQ3: Are There Differences between Groups in Distributions Humor and Offense Ratings?}

When looking at the distribution of ratings across the 6 possible values (1-5 and \textit{`I don’t get it'}) for the entire dataset (both Kaggle and Twitter), a $\chi^2$ test of homogeneity demonstrated that there were significant differences between the distributions of humor ratings between men and women ($\chi^2$= 202.25, $p <$ 0.05) and showed that women were more likely to select ‘I don’t get it’, while men were more likely to use higher ratings. We also explored if this difference translated into different average humor ratings per text and a Wilcoxon signed rank showed that men gave significantly higher ratings than women on humor ($z$ = 9684516.5, $p <$ 0.05) and the CLES score of 0.5333 indicated that men gave higher humor ratings in 53.33\% of pairs.

For general offense, a $\chi^2$ test of homogeneity showed significant differences between groups ($\chi^2$ = 430.85, $p <$ 0.05), and examining the expected versus observed counts showed that the trend seen in the humor ratings was reversed: men were more likely to choose low offense ratings and women were more likely to select higher values. In terms of averaged general offense ratings, group differences were significant ($z$ = 4260050.5, $p <$ 0.05, CLES = 0.4704), with men giving higher offense ratings in 47.04\% of pairs. 

Similarly, for personal offense, a $\chi^2$ test of was significant ($\chi^2$ = 1195.94, $p <$ 0.05) with a more pronounced trend showing that women were more likely to select a high personal offense rating, and men systematically under-selected high ratings. This led to significant differences in the average personal offense ratings per text, where men gave higher personal offense scores in only 41.5\% of pairs ($z$ = 1234096.5, $p< $0.05, CLES = 0.4146). 

When looking at age groups, a $\chi^2$ test showed significant differences in humor ratings between age groups ($\chi^2$ = 239.98, $p <$ 0.05). The oldest group, 56-70, were most likely to report ‘I don’t get it’, while annotators aged 26-40 were least likely to use this, and most likely to give high ratings. In terms of general offense, there were significant group differences ($\chi^2$ = 540.936 $p <$ 0.05), and annotators 18-40 were more likely to give lower general offense ratings, while those aged 41-70 used fewer low ratings than expected, and the group ages 56-70 was most likely to give the highest possible offense rating of 5. Group differences were more pronounced in personal offense ratings ($\chi^2$ = 1387.43, $p <$ 0.05) where the two youngest groups gave consistently lower than expected ratings of personal offense, while the older group gave consistently higher ratings. This resulted in significant differences in the average personal offense scores between groups ($\chi^2$ = 38.223, $p<$0.05).

\section{Qualitative Analysis}
\begin{figure}[h]
\caption{Relationship Between Humor and Offense by Gender}
\centering
\includegraphics[width=1.0\textwidth]{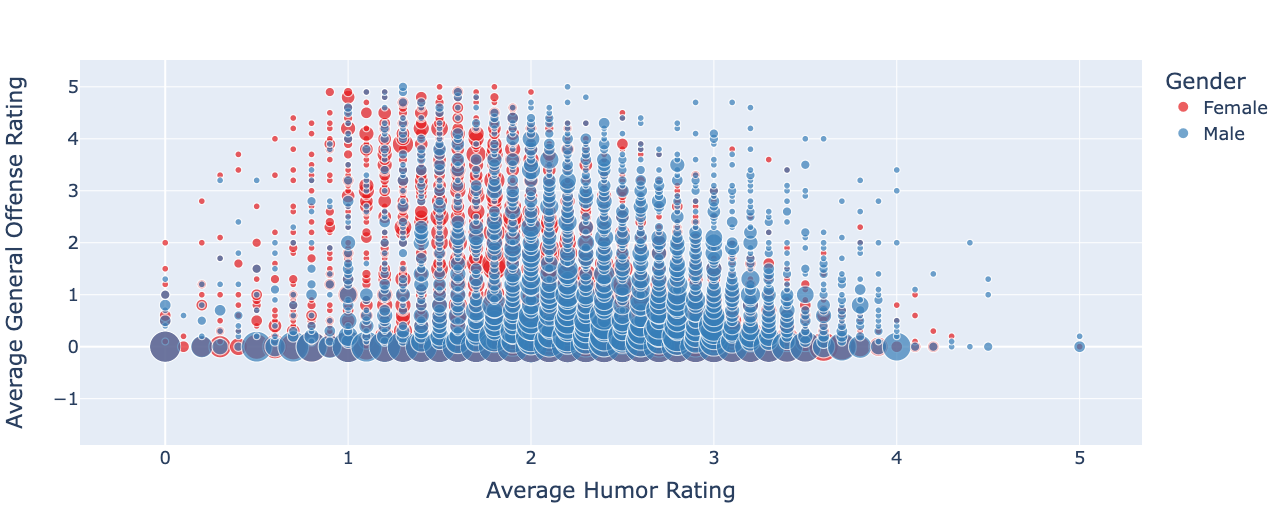}
\label{fig:humor-offense-scatter}
\end{figure}

The negative correlation for female annotators between humor and general offense, which was uncovered in the above analysis, is succinctly illustrated in Figure \ref{fig:humor-offense-scatter}. Texts which are offensive for women tend to earn a lower humor rating, while general offense is more tolerated by men. 

To examine what type of texts male and female annotators differed on with regard to general offense ratings, we selected the top 40 texts where there was at least a 1.5 point difference between the mean general offense score given by male and female annotators. We labeled the topic or target of the texts and five annotators rated whether the content was aggressive or not. Annotators were instructed that a text should be deemed aggressive if it contained violent content or used racial slurs, and inter-annotator agreement was relatively high (Fleiss’s $\kappa$ = 0.3815).

\begin{table}[]
\caption{Sample Texts Where Annotators Differed on General Offense}
\label{tab:gender-text-differences}
\begin{tabular}{lllll}
\hline
\multicolumn{1}{c}{\textbf{Text}} & \multicolumn{2}{c}{\textbf{Humor}} & \multicolumn{2}{c}{\textbf{G. Offense}} \\ \cline{2-5} 
 & \textbf{Female} & \textbf{Male} & \textbf{Female} & \textbf{Male} \\ \hline
\begin{tabular}[c]{@{}l@{}}Why are the labia on Japanese women \\ oriented sideways instead of vertically? \\ Goes better with their eyes.\end{tabular} & 1.0 & 2.2 & 4.2 & 1.3 \\ \hline
\begin{tabular}[c]{@{}l@{}}In my spare time, I help blind kids \\ I mean the verb, not the adjective\end{tabular} & 1.3 & 2.0 & 2.2 & 0.17 \\ \hline
\begin{tabular}[c]{@{}l@{}}Two condoms walk by a gay bar... \\ One says to the other, \\ "Wanna go inside and get shitfaced?"\end{tabular} & 2.6 & 1.6 & 0.85 & 2.4 \\ \hline
\begin{tabular}[c]{@{}l@{}}What did the Jewish pirate say when \\ he heard his wife died? Argh, shiva me timbers\end{tabular} & 1.6 & 1.6 & 1.0 & 2.1 \\ \hline
\end{tabular}
\end{table}

\begin{figure}[h]
\caption{Topics and Aggression where Gender Groups Disagreed on General Offense Ratings}
\centering
\includegraphics[width=1.0\textwidth]{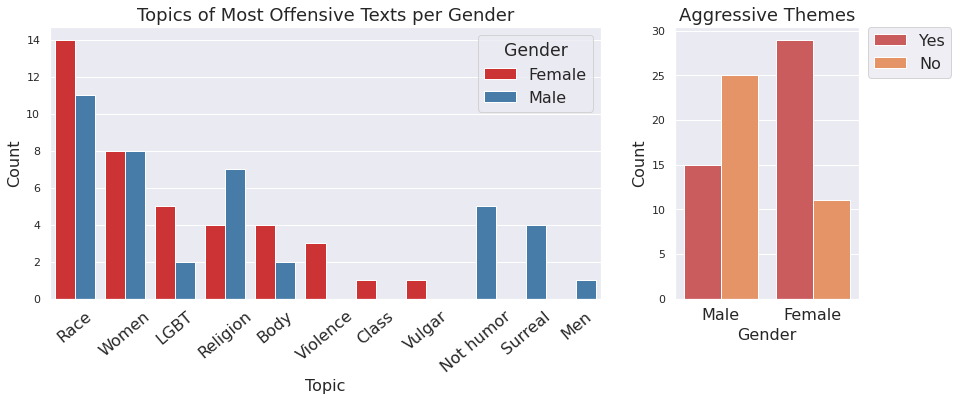}
\label{fig:gender-offense}
\end{figure}

There was a sizeable overlap of topics, with women finding texts about the LGBT community more offensive than men, while male annotators found texts about religion more generally offensive. The texts that were offensive to women tended to be aggressive, while men were more tolerant of this. Interestingly, men selected several texts which were not intended to be jokes (e.g. were drawn from accounts supporting targets of hate speech) as both humorous and offensive. 

We followed a similar procedure to examine the texts where offense ratings from different age groups differed from each other. We compared the mean general offense rating from each group to the average general offense rating from the other 3 groups combined, and looked at the top 40 texts where there was at least a 1.5 point difference. Several topics predominate, namely race, women, body (e.g. disability, body weight). The texts rated as more generally offensive by younger groups focused on these topics, but as age increased, so did the variety of topics featured. The texts selected by group 1 (the youngest group) featured more which were aggressive in nature, but as age increased, aggression was less linked to offense. 

\begin{figure}[h]
\caption{Analysis of Topics and Aggression where Age Groups Disagreed on General Offense Ratings}
\centering
\includegraphics[width=1.0\textwidth]{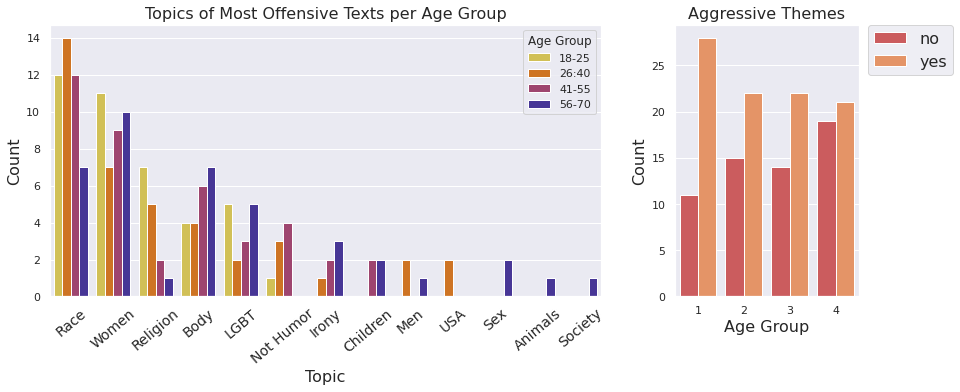}
\label{fig:age-offense}
\end{figure}

\section{Discussion}
We used a large dataset of texts rated for humor and offense, along with some demographic information about the annotators to explore differences between age and gender groups. We looked at how the groups link humor and offense, differences in humor detection and comprehension, as well as differences in the distributions of ratings. 

\textbf{RQ1}: We found that female annotators negatively link humor and offense more strongly than men. Male annotators do not link general offense with diminished humor ratings. In fact, they link humor and offense to a lesser extent, and only when personally offended. 

As regards age groups, the correlation between humor and offense was weakest in the youngest group, and grew steadily with age - as did the link between general and personal offense. 

\textbf{RQ2}: There were no differences in gender or age groups in terms of humor detection. However, when it came to humor comprehension, women selected ‘I don’t get it’ more often than men. 

\textbf{RQ3}: In terms of the distributions of ratings, women gave lower humor ratings and higher offense ratings, while men showed the opposite trend. Amongst the age groups, annotators 26-40 gave the highest ratings and the fewest reports of ‘I don’t get it’. In line with findings from RQ1, younger groups gave lower offense ratings and older groups reported higher offense. 

Some of the findings above are well attested in the humor literature, albeit in smaller-N studies. Hofmann et al.~\cite{hofmann2020gender} report that men’s tolerance of aggressive humor is one of the most consistent findings in the humor field, with seven out of eight studies mentioned replicating this result. Our qualitative work shows that in the texts on which men and women differed most on general offense, aggression featured more prominently for women. Perhaps relatedly, Proyer and Ruch~\cite{proyer2010enjoying} report that men score higher on katagelasticism - the joy of laughing at others. This may be reflected in the fact that general offense does not diminish male annotators’ humor ratings, only personal offense does.

A more surprising result is the increasingly strong negative correlation between humor and offense as age progressed. This contradicts the oft-touted idea of \textit{Generation Snowflake}, which contends that those born after 1995 tend to be the most overly reactive to offensive material~\cite{haidt2018coddling}. The older age groups - 40-55 and 56-70 - gave higher ratings of offense than their younger counterparts, and our qualitative analysis indicated that the older groups gave higher offense ratings to a wider variety of topics. 

The finding that women used the ‘I don’t get it’ label more than men is a result that may benefit from some contextualisation from the humor literature. Bell~\cite{bell2013responses} found that when shown incomprehensible jokes, women tended to explicitly state that they did not get it, while men implicitly signaled it by asking concept-checking questions. It is not possible to know whether this was the case here, but it is true that the qualitative results uncovered that men were mentioning not humorous texts as both humorous and offensive.

\subsection{Implications}
Given the gender and age group differences in ratings of humor and offense, it is evident that humor detection systems which average over all annotators' ratings fail to model the subjectivity that is inherent to this task. These systems may not generalise well on downstream tasks, such as content moderation, and may not be effective at moderating aggressive content if they are tuned to men's preferences, or alternatively may be more restrictive if tuned to women's preferences. Furthermore, as sociologists have pointed out \cite{lockyer2005beyond}, the line between humor and offense is continually under revision in most societies, therefore not only are these responses subjective, but they are a moving target. We should focus on incorporating frameworks to include demographic knowledge in our systems, which can constantly be updated to reflect society's changing definitions of humor and offense.  

\subsection{Limitations}
It is a limitation that the dataset did not afford the opportunity to explore the interaction between age and gender. As each text has approximately 20 annotations per text, splitting these into 8 groups to model age and gender would not have provided sufficient statistical power. Similarly, it is a limitation that there were insufficient annotations from gender non-conforming annotators, as there is a dearth of literature on their reactions to humor and offense.
The lack of annotators that self-identify with genders other than female and male has been noticed in the past in different tasks as well~\cite{excell2021towards,prabhakaran2021releasing}.

A final constraint is that we are modelling only one half of the humorous interaction - the recipient of the joke. Excluding the teller of the joke can deny the recipient some important context needed to enjoy the joke, and different tellers can mitigate the responses. Future work should include this dimension. 

\section{Conclusion}
We present the first analysis of the demographic data provided with the HaHackathon data - a large dataset used to train systems for computational humor detection. Our findings indicate that women negatively link humor to offense, while men only do so if they are personally offended. Links between humor and offense grew with age. There were no differences in humor detection by gender or age groups, but women and older annotators indicated that they did not understand jokes more than men. Distributions of humor and offense ratings replicated findings from humor research, namely that men gave higher humor ratings and lower offense ratings. We hope that these findings will inform future frameworks for computational humor detection and dataset creation. 

\subsection*{Acknowledgements}
This work was supported in part by the EPSRC Centre for Doctoral Training in Data Science, funded
by the UK Engineering and Physical Sciences Research Council (grant EP/L016427/1) and the University of Edinburgh.

%
%
%


\bibliographystyle{splncs04}
\bibliography{refs}

\newpage


\end{document}